\documentclass{fairastra}
\usepackage[T1]{fontenc}
\usepackage{lmodern}
\usepackage{multirow}
\usepackage{soul}
\usepackage{pifont}
\usepackage[T1]{fontenc}

\usepackage{graphicx}
\usepackage{amsmath,amssymb}

\usepackage{makecell}
\usepackage[dvipsnames]{xcolor}
\usepackage{color, colortbl}

\usepackage{caption}
\usepackage{algorithm}

\usepackage{xspace}
\makeatletter
\DeclareRobustCommand\onedot{\futurelet\@let@token\@onedot}
\def\@onedot{\ifx\@let@token.\else.\null\fi\xspace}

\makeatother

\usepackage{newtxtext}

\usepackage{algpseudocode}
\definecolor{catgray}{gray}{0.92}

\usepackage[dvipsnames]{xcolor} 
\definecolor{FutureOrange}{HTML}{EC866D}

\title{AstraNav-World: World Model for Foresight Control and Consistency}

\usepackage{marvosym}
\usepackage{fontawesome5}
\author[1,2*\dagger]{Jintao Chen}
\author[1*\raisebox{-3pt}{\textsuperscript{\Letter}}]{Junjun Hu}
\author[1,*]{Haochen Bai}
\author[1]{Minghua Luo}
\author[1,2]{Xinda Xue}
\author[1,3]{Botao Ren}
\author[1,2]{Chengyu Bai}
\author[1]{Shichao Xie}
\author[1]{Ziyi Chen}
\author[1]{Fei Liu}
\author[1]{Zedong Chu}
\author[1]{Xiaolong Wu}
\author[1]{Mu Xu}
\author[2\raisebox{-3pt}{\textsuperscript{\Letter}}]{Shanghang Zhang}

\affiliation[1]{\textbf{Amap Alibaba}}
\affiliation[2]{PKU}
\affiliation[3]{THU}

\contribution[*]{Equal Contribution}
\contribution[\dagger]{Project Lead.}
\contribution[]{\raisebox{-1pt}{\textsuperscript{\Letter}}Co-corresponding authors}

\abstract{
\begin{abstract}
 EEmbodied navigation in open, dynamic environments demands accurate foresight of how the world will evolve and how actions will unfold over time. 
 We propose AstraNav-World, an end-to-end world model that jointly reasons about future visual states and action sequences within a unified probabilistic framework. 
 Our framework integrates a diffusion-based video generator with a vision-language policy, enabling synchronized rollouts where predicted scenes and planned actions are updated simultaneously. 
Training optimizes two complementary objectives: generating action-conditioned multi-step visual predictions and deriving trajectories conditioned on those predicted visuals.
This bidirectional constraint makes visual predictions executable and keeps decisions grounded in physically consistent, task-relevant futures, mitigating cumulative errors common in decoupled "envision-then-plan" pipelines. 
Experiments across diverse embodied navigation benchmarks show improved trajectory accuracy and higher success rates. Ablations confirm the necessity of tight vision–action coupling and unified training, with either branch removal degrading both prediction quality and policy reliability. In real-world testing, AstraNav-World demonstrated exceptional zero-shot capabilities, adapting to previously unseen scenarios without any real-world fine-tuning. These results suggest that AstraNav-World captures transferable spatial understanding and planning-relevant navigation dynamics, rather than merely overfitting to simulation-specific data distribution.
Overall, by unifying foresight vision and control within a single generative model, we move closer to reliable, interpretable, and general-purpose embodied agents that operate robustly in open-ended real-world settings.
\end{abstract}
}
\date{December 25th, 2025}
\astradata[Project Page]{\url{https://astra-amap.github.io/AstraNav-World.github.io/}}

\begin{document}

\maketitle

\section{Introduction}
Embodied navigation is a core capability of embodied intelligence, aiming to enable agents to act and make decisions autonomously in complex, unknown real environments~\cite{yang2025cenavflowguidedreinforcementrefinement,zhang2025mem2ego,wang2022towards,anderson2018evaluation}. Despite significant progress in perception and instruction understanding, a key reason for navigation failures remains the lack of modeling of physical laws and temporal dynamics: small prediction biases accumulate over time, ultimately undermining the effectiveness of global planning. To achieve highly reliable real-world navigation, we advocate advancing two capability pipelines in tandem—"envision the future" and "plan the future"~\cite{finn2017deep,wu2024embodied}. The former emphasizes generating credible future visual states given actions to evaluate the model's understanding of physics and causality; the latter emphasizes task-oriented action prediction, imposing constraints on visual generation so that predicted states are closer to the reachable world.

For "plan the future", a common route is to build VLA on top of a VLM by adding an action head to achieve accurate and fast trajectory prediction~\cite{nvidia2025gr00t,wang2025trackvla}. For "envision the future", a natural path is to build a world model ~\cite{ali2025world,russell2025gaia}. Methods~\cite{bar2025navigation,zhou2025learning,yao2025navmorph} like NWM follow an "imagine-then-act" paradigm: given past images and actions, it predicts future first-person frames and uses them to judge whether the action sequence is feasible. Most of these methods use a video generation model as the base model since video generation models possess strong visual priors. Additionally, to better exploit video-generation priors, some studies use large-scale video-generation pretraining to improve model generalization~\cite{lyu2025dywa,zitkovich2023rt,cheang2024gr}, adopting a unified latent space to represent vision, language, actions, and world states for cross-modal knowledge sharing, and training on multi-tasks (e.g., video prediction plus policy learning).

Long-horizon planning is another key challenge in embodied intelligence. VLA approaches, benefiting from strong vision-language understanding, tend to have better long-horizon planning potential; in contrast, world model-based long-horizon planning has been relatively underexplored. This has led to directions that combine the two, such as WorldVLA~\cite{cen2025worldvla} and CoT-VLA~\cite{cen2025worldvla,zhang2025dreamvla}: a single autoregressive Transformer serves both as a "world model" (generating future images) and an "action model" (generating control sequences). Among them, CoT-VLA emphasizes explicit reasoning via "intermediate visual states" to improve decision-making for complex instructions; WorldVLA points out that multi-step prediction errors can easily trigger planning collapse, requiring physical future consistency. Overall, VLAs focus on complex instructions and long-horizon planning, while world models focus on environment understanding and dynamics prediction; combining them produces complementary benefits.

However, most existing methods still follow a loosely coupled "envision-then-plan" paradigm: envisioning and planning are decoupled in the pipeline, which tends to amplify physical uncertainty and causal ambiguity, leading to inconsistencies between the two. To address this, we propose a unified generative navigation world model that tightly binds "envision the future" and "plan the future" within a single probabilistic framework. Building on a powerful VLM with long-horizon perception and instruction understanding~\cite{bai2025qwen2}, we deeply couple a DiT-based video generator~\cite{wan2025wan} with an action policy~\cite{xue2025omninav}, jointly modeling multi-step future visual frames and action sequences conditioned on the current state and language instructions. We optimize the two complementary objectives in parallel, and they progress in lockstep via synchronized rollout. This bidirectional constraint both provides executable visual evidence to support the rationality of planning and uses policy choices to inversely constrain visual generation to stay aligned with task intent~\cite{xiu2025egotwin}. 
As a result, we construct a long-horizon navigation architecture that is visually coherent and behaviorally effective, reducing error propagation and pipeline inconsistencies, improving robustness and success rates. Extensive experiments show that our joint framework significantly outperforms existing methods in both future trajectory prediction accuracy and overall navigation success rate. Ablation studies further validate the necessity of bidirectional coupling and joint training: removing either branch weakens the grasp of environmental physics and task constraints, causing synchronized degradation in prediction performance. This study provides three primary contributions:
\begin{itemize}
\item A unified generative framework in which future visual frames and action sequences are modeled simultaneously, and through bidirectional constraints and synchronized rollout under joint optimization, physical and causal consistency are reinforced, leveraging executable visual evidence to support the action, and generated frames are constrained by the action to maintain persistent alignment with task goals.

% \item We propose AstraNav-World, a world model architecture centered around a VLM as a central planner. This VLM is conditioned on historical and current multi-view frames alongside a language instruction. The planner then guides two distinct components: a state predictor, implemented by a VLM-conditioned video generator, and a Diffusion Policy. Crucially, our Video Generator and Policy Model are designed to simultaneously predict and generate future frames and actions, respectively, based on the vision-language embeddings generated by the central planner.
\item A world-model-based navigation architecture (AstraNav-World) is proposed, where the vlm serves as a global planner and encodes the language instruction together with historical and current multi-view observations into vision–language embeddings, which are then used to condition (i) a diffusion-based video generator for state prediction and (ii) a policy head for trajectory prediction.

\item Performance, interpretability, and zero-shot transfer: AstraNav-World achieves strong gains on multiple navigation benchmarks, improving SR/SPL and reducing NE compared to prior methods. Importantly, without any real-world fine-tuning, AstraNav-World demonstrates robust zero-shot transfer on a physical robot in previously unseen real environments. 
%Ablation studies further show that bidirectional vision–action coupling and joint training are key to these improvements.
\end{itemize}

In summary, we place "envision the future" and "plan the future" within a single generative model. Leveraging the long-horizon understanding of a strong VLM, we use physically consistent video generation and task-aligned policy prediction to obtain a coherent and long-horizon-appropriate navigation plan. This not only improves robustness and success in real environments but also makes decisions more interpretable and extensible.

\section{Related Works}

\subsection{World Model}
World models aim to capture environment dynamics and predict future states to support perception and decision making. Beyond short-horizon rollouts, recent work treats them as generators that synthesize future videos to provide visual evidence for planning. LaDi-WM~\cite{huang2025ladi} employs diffusion modeling to predict future states within a geometric-semantic latent space, enabling more consistent and generalizable state forecasting compared to pixel-based approaches. MoWM~\cite{shang2025mowm} enhances decision robustness by fusing latent and pixel predictions through feature modulation in a mixture-of-world-models architecture during planning. 

In addition, world models are increasingly employed not only for prediction but as generative, interactive environments. Interactive world models condition on actions, language, or goal images to generate controllable, step-by-step futures, showcasing strong capabilities in environment simulation. Genie series~\cite{bruce2024genie,parker2024genie} show that internet-scale video plus latent-action Transformers can yield a single, prompt-controllable world model for 2D/3D games and robotics, enabling zero-shot policy imagination without real-world rewards or labels. Yan~\cite{team2025yan} enables real-time, infinite interactive video generation via cross-domain prompts using a dynamics–rendering disentanglement. World models play the same role in the navigation domain as well, are used for scene understanding and future-frame imagination to guide planning, which provides visual evidence for exploration ~\cite{bar2025navigation}.

In terms of architecture, for video generation, autoregressive (AR) models and diffusion models have become the two dominant paradigms. AR world models serialize and autoregressively model multiple modalities—video, actions, and language—with representative work including MineWorld~\cite{guo2025mineworld}. Their advantages include natural suitability for sequence modeling, strong scalability, and effective cross-modal fusion. Diffusion-based world models (e.g., Matrix-Game 2.0~\cite{he2025matrix} and UWM~\cite{zhu2025unified}) deliver higher generation quality, and have become the preferred choice in world generation due to their superior performance. Given our focus on generating high-fidelity, physically consistent future visual evidence for navigation and on leveraging the world understanding and physical priors from large-scale pretraining, we adopt a diffusion-based world model as our core architecture.

\subsection{Combining world models with VLA}

World models excel at physically consistent short-horizon rollouts for near-field planning and closed-loop control, while VLA models capture high-level semantics and generate coherent long-horizon actions. Recent work exploits this complementarity by tightly coupling the two, yielding more reliable long-horizon planning and greater interpretability.

WorldVLA~\cite{cen2025worldvla} integrates a latent world model with a VLA and produces a visual chain of thought (V-CoT) during inference to aid decision-making and enhance interpretability. DreamVLA~\cite{zhang2025dreamvla} adopts a Dreamer-style latent world model to perform “dream-like” visual rollouts and incorporates intermediate multimodal representations—such as dynamic maps, depth maps, and semantic maps into the reasoning process, providing visual evidence for multi-step planning and manipulation. CoT-VLA~\cite{zhao2025cot} explicitly injects visual intermediate goals into VLA by first autoregressively predicting future frames as visual subgoals and then generating short action sequences, thereby unifying temporal planning and interpretability. GigaBrain-0~\cite{team2025gigabrain} integrates RGB-D sensory modeling with embodied chain-of-thought supervision to systematically enhance reasoning over spatial geometry, object-state estimation, and long-horizon temporal dependencies. FlowVLA~\cite{zhong2025flowvla} uses Visual CoT to predict optical flow before next-frame synthesis, aligning dynamics pretraining with action generation and improving physical plausibility and sample efficiency.

Unlike the above works, we move away from the loosely coupled "envision-then-then" paradigm and instead deeply couple video generation (world modeling) and action generation (policy modeling) within a single probabilistic generative process with synchronized rollout.  At each step, the two are mutually conditioned and cross-corrected, reducing cross-modal inconsistencies and long-horizon error accumulation.

\begin{figure*}[!t]
    \centering
    \includegraphics[width=\linewidth]{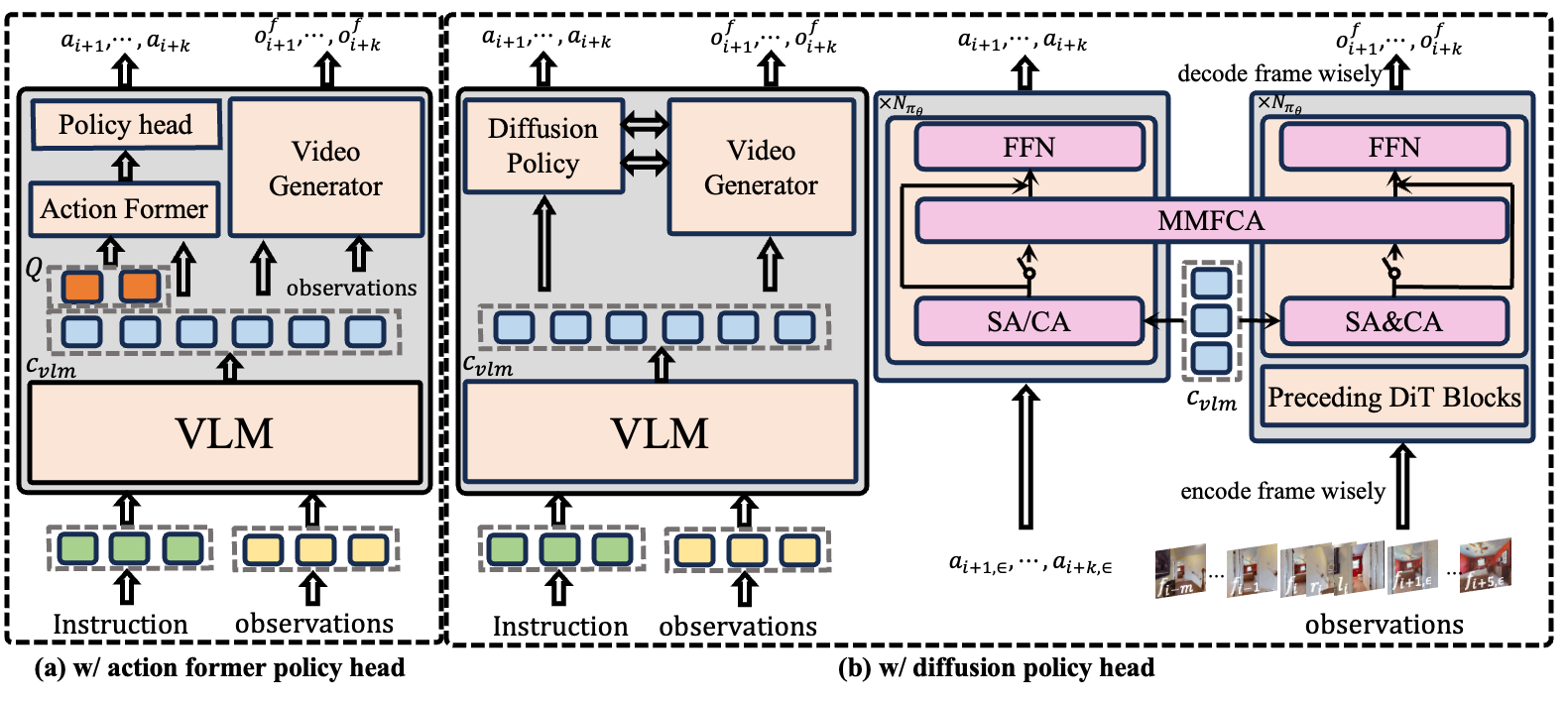}
    \caption{Overview of AstraNav World architecture. Our model adapts to two different policy heads within a unified framework: (a) a traditional policy head with an Action Former for direct action prediction, and (b) a diffusion policy head for probabilistic action generation. These two policy streams share a common VLM planner ($\tau_{\theta}$), which processes instructions and visual history to generate high-level conditional tokens. The architecture also includes a VLM conditional video generator ($\upsilon_{\theta} $) for predicting and planning consistent future visual scenes. When using diffusion strategy flow, multimodal fusion cross attention (MMFCA) can be selectively used in overlapping DiTBlocks to interconnect, achieving bidirectional information flow between action and visual prediction.}
    \label{fig:overview}
\end{figure*}

\section{Method}

\subsection{Overall Architecture}

Our AstraNav-World is a unified generative world model that tightly couples future vision prediction with action prediction within a single probabilistic framework. Unlike conventional "envision-then-plan" approaches that treat these components as separate modules, our architecture ensures bidirectional consistency between visual predictions and action sequences through joint modeling.

As illustrated in Figure~\ref{fig:overview}, AstraNav-World consists of three core components:
\begin{itemize}
    \item \textbf{Vision-Language Model (VLM) Planner}: Serves as the high-level reasoning module that processes instructions and visual history to generate semantic guidance.
    \item \textbf{Video Generator}: Predicts future visual observations based on the VLM's contextual understanding.
    \item \textbf{Action Policy Head}: Generates future action sequences, implemented through dual policy methods (transformer-based and diffusion-based).
\end{itemize}

The key innovation is that these components are trained end-to-end with bidirectional constraints, ensuring that:
\begin{itemize}
    \item VLM understands instructions and makes macro planning.
    \item Action sequences are grounded in visually plausible futures.
    \item Visual predictions are consistent with executable actions.
\end{itemize}

This unified approach eliminates the error accumulation inherent in sequential prediction-planning pipelines, while enabling long-horizon planning with visual foresight. Our model is implemented using Qwen-2.5-VL-3B as the VLM backbone and Wan2.2-TI2V-5B as the foundation for the video generation and policy components.

\subsection{VLM as Central Planner}
% \subsection{Central Planner: VLM}
The VLM, denoted as $\tau_{\theta}$, serves as the high-level reasoning core of our agent. At each timestep, it takes as input the natural language instruction $I$ and a sequence of historical visual observations $\mathcal{O}_{\text{hist}} = \{o_{i-k}, \dots, o_i\}$. 

\paragraph{\textbf{Output Representation}}
The VLM generates a set of \textbf{Vision-Language Embeddings}, $C \in \mathbb{R}^{L \times D}$ where $L$ is the sequence length and $D=2048$ is the embedding dimension. These embeddings contain two key components:
\begin{itemize}
    \item \textbf{Goal-oriented Semantic Features}: Encoded from the instruction, capturing the high-level navigation objective.
    \item \textbf{Spatial Context Features}: Encoding the historical and current visual semantic and spatial features.
\end{itemize}

These embeddings provide a unified high-level directive for both the action and vision prediction streams, ensuring cohesive planning and prediction. The key is that this representation not only enables the model to maintain a holistic understanding of long-horizon tasks, but also allows it to flexibly adapt to immediate changes in the environment.
 
\subsection{VLM-Conditioned Video Generator}
\subsubsection{\textbf{Basic Architecture}}
Our approach leverages Wan-2.2-TI2V-5B ~\cite{wan2025wan}, a bidirectional diffusion-based video generation system initially trained for short video synthesis from text or image inputs. At its core, the architecture integrates two key components: a spatio-temporal variational autoencoder(ST-VAE) and diffusion transformer networks (DiTs). The ST-VAE features a 3D causal structure that efficiently compresses high-resolution video content into a compact latent representation, achieving a 16$\times$ spatial compression ratio and a 4 $\times$ temporal compression ratio. 
The diffusion process is implemented through a series of diffusion transformers (DiTs) that include patchifying and unpatchifying layers along with 30 transformer blocks. In the original Wan architecture, input text descriptions are encoded with umT5~\cite{chung2023unimax} and integrated via cross-attention, while denoising timestep embeddings are injected through a shared MLP module. For I2V tasks, the model will concatenate the latent of the input image and noisy latent in the temporal dimension and set corresponding timestep of first frame latent to zero. In our task, we concatenate all observations in the batch dimension instead of the original temporal dimension to avoid temporal compression of ST-VAE and we set corresponding timestep of historical and current latent frames to zero.

\subsubsection{\textbf{VLM planner as Conditional Encoder.}}
To ensure the predicted future observations are aligned with the VLM's high-level plan, we replace the conventional text encoder of the video model with our VLM planner, $\tau_{\theta}$. The Vision-Language Embeddings from $\tau_{\theta}$ are injected into the video diffusion model's DiT architecture via cross-attention layers. This allows the VLM to directly steer the generation process, ensuring the synthesized future frames are semantically consistent with VLM's planning.

\subsubsection{\textbf{3D-RoPE Rearrangement.}}
Following the original settings, our framework employs 3D Rotary Position Embeddings~\cite{su2024roformer} in DiT blocks. To handle the combined input of historical frames $o_{i-k}, \dots, o_{i-1}$ (with natural temporal indexing $t_j = j$) and current multi-view observations, we introduce a 3D-RoPE rearrangement strategy. The historical frames maintain standard temporal ordering, the current left, front, and right views at $t_i$ are virtually arranged side-by-side along the width axis, preserving shared temporal and height indices while encoding their spatial-temporal relationships. For a frame with original width $W$, the coordinate transformation works as follows:

\begin{align}
PE_{t_i}^{\text{front}}  &\mapsto (t', h', w') = (t_i, h, w) \\
PE_{t_i}^{\text{right}}    &\mapsto (t', h', w') = (t_i, h, w + W) \\
PE_{t_i}^{\text{left}}     &\mapsto (t', h', w') = (t_i, h, w + 2W)
\end{align}

\subsubsection{\textbf{Optimization Objective.}}

The video generator is trained using a conditional denoising objective specifically designed for navigation context prediction. During training, we apply a differential noise scheduling strategy to distinguish between observed and predicted frames. The historical front perspective frames $o^f_{i-m}, \dots, o^f_{i-1}$ and the current multi-view observations ($o^f_{i}, o^l_{i}, o^r_{i}$) are provided with a minimal noise level ($\sigma_{\text{obs}} \approx 0.05$), effectively treating them as clean conditioning inputs. In contrast, the future front perspective frames $o^f_{i+1}, \dots, o^f_{i+N}$ are trained with noise added by Flow Matching method.

Let $z^{\text{future}} = \mathcal{E}(o^{\text{future}})$ ($\mathcal{E}$ means the encoder of ST-VAE) denote the sequence of latent representations for the ground-truth future frames. For each training iteration, we sample a noise level $\sigma_t$ and a noise sample $\epsilon \sim \mathcal{N}(0, I)$. The latent state at time $t$ is constructed as $z_t = (1 - t) \cdot z^{\text{future}} + t \cdot \epsilon$, while the conditioning frames (historical and current) remain at a near-zero noise level. The target velocity field is defined as $u_t = \epsilon - z^{\text{future}}$.

The optimization objective focuses exclusively on the future frames, formulated as:
\begin{equation}
\mathcal{L}_{\text{VG}} = \mathbb{E}_{t, z^{\text{future}}, C} \left[ \left\| v_{\theta}(z_t, t, C) - (\epsilon - z^{\text{future}}) \right\|^2 \right]
\end{equation}
where $v_{\theta}$ represents the video generation network, $C = \{c_{\text{vlm}}\}$ contains the VLM planner's contextual embeddings, and the expectation is taken only over the future frames.

\subsection{Action Policy Head}

\subsubsection{Action Former Policy}
The first policy implementation employs a query-based Transformer architecture that processes the VLM's contextual embeddings to generate action sequences. This approach utilizes a fixed set of $N_q$ learnable query vectors $Q \in \mathbb{R}^{N_q \times D}$ that are initialized randomly and updated during training. These queries interact with the VLM's output embeddings $E_{\text{vlm}} \in \mathbb{R}^{L \times D}$ through a multi-layer Transformer encoder, consisting of $N_{\text{enc}}$ stacked blocks with multi-head self-attention and feed-forward networks. Then the queries can capture temporal dependencies and contextual relationships and produce refined query representations $Q' \in \mathbb{R}^{N_q \times D}$. These representations are then passed through a multi-layer perceptron (MLP) head that maps each query to a step of action in the navigation space, yielding the predicted action sequence $A = \{a_{i+1}, \dots, a_{i+N}\}$. This architecture provides an efficient mechanism for deterministic action generation while maintaining computational efficiency.

The Action Former policy generates actions represented as $A = (X, Y, \cos(\theta), \sin(\theta), \alpha)$, where $(X, Y)$ denote the relative position displacement, $(\cos(\theta), \sin(\theta))$ encode the heading angle in a rotationally invariant manner, and $\alpha$ is a binary flag indicating arrival at the target.

The loss function is composed of three distinct components that address different aspects of the navigation task:

\begin{itemize}
    \item \textbf{Position Loss}: For the $(X, Y)$ components, we apply L1 loss to measure the absolute error between predicted and ground-truth displacements:
    \begin{equation}
    \mathcal{L}_{\text{pos}} = \frac{1}{N} \sum_{n=1}^{N} \left( |X_n - X_n^*| + |Y_n - Y_n^*| \right)
    \end{equation}
    
    \item \textbf{Angle Loss}: The heading angle is optimized using cosine similarity between the predicted and ground-truth direction vectors:
    \begin{equation}
    \mathcal{L}_{\text{angle}} = 1 - \frac{1}{N} \sum_{n=1}^{N} \left( \cos(\theta_n)\cos(\theta_n^*) + \sin(\theta_n)\sin(\theta_n^*) \right)
    \end{equation}
    This formulation ensures rotational consistency and avoids the discontinuity issues that would occur with direct angle regression.
    
    \item \textbf{Arrival Loss}: The arrival flag is optimized using binary cross-entropy with logits:
    \begin{equation}
    \mathcal{L}_{\text{arrive}} = -\frac{1}{N} \sum_{n=1}^{N} \left[\alpha_n^* \log(\sigma(\alpha_n)) + (1-\alpha_n^*) \log(1-\sigma(\alpha_n)) \right]
    \end{equation}
    where $\sigma$ denotes the sigmoid function.
\end{itemize}

The total loss for the Action Former policy is a weighted combination:
\begin{equation}
\mathcal{L}_{\text{PH}} = \lambda_1 \mathcal{L}_{\text{pos}} + \lambda_2 \mathcal{L}_{\text{angle}} + \lambda_3 \mathcal{L}_{\text{arrive}}
\end{equation}
with $\lambda_1=1.0$, $\lambda_2=1.0$, and $\lambda_3=1.0$ empirically determined to balance the contributions of each component. This multi-objective formulation ensures that the policy simultaneously optimizes for accurate position prediction, correct orientation, and appropriate termination conditions.

\subsubsection{Diffusion Policy}
The second policy implementation adopts a diffusion-based approach that generates actions through a denoising process, providing probabilistic action predictions that better capture the uncertainty inherent in navigation tasks. Each block contains one attention and one FFN, with self-attention and cross-attention alternately. The diffusion policy operates on noisy action sequences $A_t$ and gradually refines them to produce clean, executable trajectories. A key innovation in this implementation is the integration of the Multimodal Fusion Cross-Attention (MMFCA) module between the last 8 overlapping blocks (with cross attention) of the diffusion policy and video generator streams. The MMFCA enables bidirectional information flow through dual cross-attention operations:

\begin{itemize}
    \item \textbf{Action-to-Video Attention}: The refined action representations serve as queries ($Q_A$) that attend to the video latent representations ($K_V, V_V$), ensuring actions are grounded in visually plausible futures
    \item \textbf{Video-to-Action Attention}: The video latent representations serve as queries ($Q_V$) that attend to the action representations ($K_A, V_A$), ensuring visual predictions remain causally consistent with the planned actions
\end{itemize}

The MMFCA module is controlled by a binary switch $\gamma \in \{0,1\}$. When $\gamma=1$ we enable bidirectional fusion between the video generator and the policy via MMFCA module, so that video prediction and action generation are coupled and can be produced synchronously within the same rollout. When $\gamma=0$ the fusion is disabled and the two streams operate independently; in particular, for efficient inference, we can skip the video generator entirely and run only the policy to predict actions, which substantially reduces computation.

The Diffusion Policy is trained using Flow Matching method ~\cite{lipman2022flow, liu2022flow}, a continuous-time diffusion framework that learns to predict the velocity field that transports noise to the target distribution. Given a ground-truth action sequence $A_{\text{future}}$ and a noise sample $\epsilon \sim \mathcal{N}(0, I)$, we construct a linear interpolation path $A_t = (1-t) \cdot A_{\text{future}} + t \cdot \epsilon$. The target velocity field is defined as $v_t =  \epsilon - A_{\text{future}}$.

The loss function for the Diffusion Policy is formulated as:
\begin{equation}
\mathcal{L}_{\text{PH}} = \mathbb{E}_{t, A_{\text{future}}, \epsilon, C} \left[ \left\| v_{\phi, \theta}(A_t, t, C) - (\epsilon - A_{\text{future}}) \right\|^2 \right]
\end{equation}
where $v_{\phi, \theta}$ denotes the velocity prediction network, and $C$ represents the VLM planner's contextual embeddings.

\subsection{Speed Optimization}

The intrinsic speed of video generation, particularly with diffusion transformer models , can be a bottleneck for real-time navigation tasks that demand high-frequency actions. To address this, we propose \textbf{Sparse Foresight Scheduling} (SFS). Through joint modeling and joint training, the VLM possesses sufficiently strong planning and generalization capabilities, extending these capabilities to the policy level. So Instead of generating a future frame and action at every single time step, we perform this joint generation process at fixed, pre-determined intervals. This is a pragmatic decision, as many navigation scenarios involve prolonged periods of simple, consistent behaviors such as straight-line movement, which do not necessitate instantaneous updates to the world model. By generating future predictions only at these sampled intervals, we significantly reduce computational overhead and save valuable time.

During inference, our system adapts its computational strategy based on the selected policy head. When utilizing the Action Former policy head, we completely deactivate the video generator, relying solely on the query-based Transformer architecture to generate action sequences without visual prediction.

%This configuration provides maximum computational efficiency for scenarios where visual foresight is not required.
Because we disabled MMFCA for 50\% of the training process, our diffusion policy has sufficient independence to allow us to intermittently disable joint prediction and only predict future actions. For the Diffusion Policy head, we activate the video generator only at fixed interval steps (every 10 steps in our implementation), while keeping it deactivated during intermediate steps. This selective activation approach allows our model to benefit from visual foresight when it matters most while significantly reducing computational overhead, achieving an optimal trade-off between predictive accuracy and inference speed.
\section{Experiments}
\subsection{Data Process}
Our data resources encompass two types of navigation tasks: instruction-goal navigation and object-goal navigation.

\textbf{Instruction-goal navigation data}. Under Habitat’s continuous environment configuration (VLN-CE) and Matterport3D indoor scenes~\cite{chang2017matterport3d}, we construct navigation episodes based on the public instruction–path pairs from R2R~\cite{anderson2018vision} and RxR~\cite{ku2020room}. For each trajectory, we store multimodal information, including a panoramic sequence composed of three camera views (left, front, right), the corresponding sequences of discrete actions and continuous waypoints, and the paired natural-language instruction.

\textbf{Object-goal navigation data}. Following the open-vocabulary object navigation setting (OVON)~\cite{yokoyama2024hm3d}, we randomly sample (start pose, target object category) pairs in HM3D scenes~\cite{ramakrishnan2021habitat}. A built-in shortest-path planner guides the agent to the vicinity of the target to generate samples, and we record each episode in the same format as the instruction-goal data (three-view panoramic frames, discrete action and continuous waypoint sequences, and the associated textual description).

\textbf{Navigation data Process}. Navigation comprises four discrete action types (forward, left, right, stop). Each action step corresponds to a continuous trajectory point composed of position and orientation: position is a 3D coordinate ($X$, $Y$, $Z$), and rotation is a quaternion ($w$, $x$, $y$, $z$). During trajectory prediction, we first convert each 6-DoF pose to a local coordinate frame: the current front-view frame serves as the origin and heading reference, and all other trajectory points are transformed relative to this reference. We then project the 3D pose onto the ground plane, retaining ($X$, $Y$) and the heading angle $\theta$, and add an "arrival" flag $\alpha$ at each step. Finally, each trajectory point (action) is represented as a quadruple ($X$,$Y$,$\theta$,$\alpha$), namely $A$=($X$,$Y$,$\theta$,$\alpha$), which serves as the primary action optimization target.

\subsection{Implementation Details}
% \paragraph{Training Setup}
For the Vision-Language Model, we utilize \textbf{Qwen2.5-VL-3B} and perform full-parameter supervised fine-tuning. The video generator is based on \textbf{Wan2.2-TI2V-5B} and is fine-tuned using the LoRA method with both rank and scale set to $128$. Empirical observations indicate that LoRA training offers superior stability and convergence speed for the video module.
For the policy components, we implement two distinct variants: 
(i) The \textbf{Action Former} variant employs $5$ learnable query embeddings processed through $4$ stacked Transformer blocks to predict deterministic trajectories. 
(ii) The \textbf{Diffusion Policy} variant utilizes a deeper architecture with $16$ Transformer layers. Furthermore, we integrate the MMFCA module between the Diffusion Policy and the final $8$ blocks (with cross attention) of the video generator to facilitate deep cross-modal interactions. 
We set the learning rate to $1\times 10^{-5}$ and adopt a cosine decay schedule. 
All experiments are conducted on a cluster equipped with $96$ \texttt{H20} GPUs.

% \subsection{Training de}
\subsubsection{VLM pretraining}
To endow our central planner with broad world knowledge and robust reasoning capabilities, we initialize the Vision-Language Model (VLM), $\tau_{\theta}$, from a checkpoint pretrained on a diverse mixture of large-scale, general-purpose vision-language datasets. These datasets cover fundamental tasks such as Visual Question Answering (VQA) and object grounding, which provide the model with a foundational understanding of object semantics and spatial relationships and instruction comprehension. In addition, the navigation dataset of R2R, RxR and Ovon as described above are also added to pretrain the VLM in an autoregressive way following the work of OmniNav~\cite{xue2025omninav}. This pretraining process ensures that the VLM is not only a knowledgeable reasoner but also an expert planner adapted to the visual features and action-oriented instructions.

\subsubsection{Video-Action Prediction Training}

The training of AstraNav-World follows a two-stage training strategy designed to effectively leverage pre-trained components while optimizing the complex interactions between visual and action prediction.

\paragraph{\textbf{Total loss}}
The overall training objective combines the video generator loss ($\mathcal{L}_{\text{VG}}$) and policy head loss ($\mathcal{L}_{\text{PH}}$) into a unified framework:
\begin{equation}
\mathcal{L}_{\text{Total}} = \mathcal{L}_{\text{VG}} + \lambda \mathcal{L}_{\text{PH}}
\end{equation}
where $\lambda = 1.0$ is the balancing coefficient empirically determined to optimize the trade-off between visual prediction accuracy and action prediction quality.

\paragraph{\textbf{Training Strategy}}
Our training process consists of two carefully designed stages:

\textbf{Stage 1: Component-Specific Pretraining.}
The VLM planner is initialized from a checkpoint pretrained on large-scale vision-language datasets and kept frozen during this entire stage. We first train the video generator independently using $\mathcal{L}_{\text{VG}}$, with the Flow Matching objective focusing exclusively on future frame prediction. Subsequently, we train the policy head independently using $\mathcal{L}_{\text{PH}}$, optimizing for both position accuracy and action semantics. This stage provides strong initializations for both components while avoiding early optimization conflicts and ensuring each module develops its core capabilities before integration.

\textbf{Stage 2: Joint Fine-tuning.}
In this stage, we unfreeze all components and jointly optimize the entire model with $\mathcal{L}_{\text{Total}}$. For models using the Diffusion Policy head, we randomly enable the MMFCA module with 50\% probability during training, which ensures the policy develops robust standalone capabilities while maintaining the ability to leverage visual guidance when available. This selective activation strategy prevents over-reliance on visual feedback and guarantees the diffusion policy can operate effectively even when visual information is limited or when the video generator is deactivated during inference for computational efficiency.

\begin{table}[t] 
\centering
\caption{Main comparison with prior methods on the Val-Unseen split of R2R-CE and RxR-CE (vg indicates video-generator,\\ * indicates methods using the waypoint predictor).}
\label{tab:r2r_rxr}
\setlength{\tabcolsep}{2pt}
\scriptsize                

\resizebox{\linewidth}{!}{
\begin{tabular}{lcccccccccccc}
\toprule
\textbf{Method} & \multicolumn{4}{c}{\textbf{Observation}} & \multicolumn{4}{c}{\textbf{R2R-CE Val-Unseen}} & \multicolumn{3}{c}{\textbf{RxR-CE Val-Unseen}} \\
\cmidrule(lr){2-5} \cmidrule(lr){6-9} \cmidrule(lr){10-12}
& S.RGB & Pano. & Depth & Odo. & NE$\downarrow$ & OS$\uparrow$ & SR$\uparrow$ & SPL$\uparrow$ & NE$\downarrow$ & SR$\uparrow$ & SPL$\uparrow$ \\
\midrule
HPN+DN*~\citep{krantz2021waypoint} & & \checkmark & \checkmark & \checkmark & 6.31 & 40.0 & 36.0 & 34.0 & - & - & - \\
CMA*~\citep{hong2022bridging} & & \checkmark & \checkmark & \checkmark & 6.20 & 52.0 & 41.0 & 36.0 & 8.76 & 26.5 & 22.1 \\
Sim2Sim*~\citep{krantz2022sim} & & \checkmark & \checkmark & \checkmark & 6.07 & 52.0 & 43.0 & 36.0 & - & - & - \\
GridMM*~\citep{wang2023gridmm} & & \checkmark & \checkmark & \checkmark & 5.11 & 61.0 & 49.0 & 41.0 & - & - & - \\
DreamWalker*~\citep{wang2023dreamwalker} & & \checkmark & \checkmark & \checkmark & 5.53 & 59.0 & 49.0 & 44.0 & - & - & - \\
Reborn*~\citep{an20221st} & & \checkmark &\checkmark & \checkmark & 5.40 & 57.0 & 50.0 & 46.0 & 5.98 & 48.6 & 42.0 \\
ETPNav*~\citep{an2024etpnav} & & \checkmark & \checkmark & \checkmark & 4.71 & 65.0 & 57.0 & 49.0 & 5.64 & 54.7 & 44.8 \\
HNR*~\citep{wang2024lookahead} & & \checkmark & \checkmark & \checkmark & 4.42 & 67.0 & 61.0 & 51.0 & 5.50 & 56.3 & 46.7 \\
\midrule
AG-CMTP~\citep{chen2021topological} & & \checkmark & \checkmark & \checkmark & 7.90 & 39.0 & 23.0 & 19.0 & - & - & - \\
R2R-CMTP~\citep{chen2021topological} & & \checkmark & \checkmark & \checkmark & 7.90 & 38.0 & 26.0 & 22.0 & - & - & - \\
InstructNav~\citep{long2024instructnav} & & \checkmark & \checkmark & \checkmark & 6.89 & - & 31.0 & 24.0 & - & - & - \\
LAW~\citep{raychaudhuri2021language} & \checkmark & & \checkmark & \checkmark & 6.83 & 44.0 & 35.0 & 31.0 & 10.90 & 8.0 & 8.0 \\
CM2~\citep{georgakis2022cross} & \checkmark & & \checkmark & \checkmark & 7.02 & 41.0 & 34.0 & 27.0 & - & - & - \\
WS-MGMap~\citep{chen2022weakly} & \checkmark & & \checkmark & \checkmark & 6.28 & 47.0 & 38.0 & 34.0 & - & - & - \\
AO-Planner~\citep{chen2025affordances} & & \checkmark & \checkmark & & 5.55 & 59.0 & 47.0 & 33.0 & 7.06 & 43.3 & 30.5 \\
Seq2Seq~\citep{krantz2020beyond} & \checkmark & & \checkmark & & 7.77 & 37.0 & 25.0 & 22.0 & 12.10 & 13.9 & 11.9 \\
CMA~\citep{krantz2020beyond} & \checkmark & & \checkmark & & 7.37 & 40.0 & 32.0 & 30.0 & - & - & - \\
NaVid~\citep{zhang2024navid} & \checkmark & & & & 5.47 & 49.0 & 37.0 & 35.0 & - & - & -  \\
Uni-NaVid~\citep{zhang2024uni} & \checkmark & & & & 5.58 & 53.5 & 47.0 & 42.7 & 6.24 & 48.7 & 40.9  \\
NaVILA~\citep{cheng2024navila} & \checkmark & & & & 5.22 & 62.5 & 54.0 & 49.0 & 6.77 & 49.3 & 44.0 \\
StreamVLN~\citep{wei2025streamvln} & \checkmark & & & & 4.98 & 64.2 & 56.9 & 51.9 & 6.22 & 52.9 & 46.0 \\
CorrectNav~\citep{yu2025correctnav} & \checkmark & & & & 4.24 & 67.5 & 65.1 & 62.3 & 4.09 & 69.3 & 63.3 \\
\midrule

\textbf{AstraNav-World} w/ Action Former & &\checkmark& & & 3.93 & 73.1 & 67.2 & 64.2 & 3.93 & 70.4 & 59.6 \\
\textbf{AstraNav-World} w/ Diffusion Policy & &\checkmark& & & \textbf{3.86} & \textbf{73.9} & \textbf{67.9} & \textbf{65.4} & \textbf{3.82} & \textbf{72.9} & \textbf{61.5} \\
\bottomrule
\end{tabular}
}% end resizebox
\end{table}

\begin{table}[h]
  \centering
  \captionsetup{justification=centering, singlelinecheck=false}
  \caption{Evaluation of object-goal navigation on HM3D-OVON.}
  \label{tab:ovon}
  \small
  \begin{tabular}{lcc}
    \toprule
    \textbf{Method} & \multicolumn{2}{c}{\textbf{Val-Unseen}} \\
    \cmidrule(lr){2-3}
    & \textbf{SR}$\uparrow$ & \textbf{SPL}$\uparrow$ \\
    \midrule
    VLFM~\citep{yokoyama2024vlfm}        & 35.2 & 19.6 \\
    DAgRL+OD~\citep{yokoyama2024hm3d}     & 37.1 & 19.8 \\
    Uni-NaVid~\citep{zhang2024uni}       & 39.5 & 19.8 \\
    MTU3D~\citep{zhu2025move}            & 40.8 & 12.1 \\   
    \midrule
    \textbf{AstraNav-World} w/ Action Former                & 45.1 & 28.3 \\
    \textbf{AstraNav-World} w/ Diffusion Policy              & \textbf{45.7} & \textbf{28.7} \\  
    \bottomrule
  \end{tabular}
\end{table}

\subsection{Main Results}
\subsubsection{Metric Comparison} 
We assess navigation performance using four standard metrics: success rate (SR), oracle success (OS), success weighted by path length (SPL), and navigation error (NE). Our evaluation setup mirrors established practice in the literature and follows the protocols used in prior work~\cite{zhang2024uni,zhu2025move}. Across instruction-following benchmarks R2R-CE and RxR-CE, AstraNav-World demonstrates significant improvements with both policy variants. The Action Former policy achieves a 2.1\% absolute improvement in success rate over prior methods on R2R-CE and 1.1\% on RxR-CE. The Diffusion Policy with MMFCA further enhances performance, delivering an additional 0.7\% gain on R2R-CE and 2.5\% on RxR-CE compared to the Action Former variant (see Tables \ref{tab:r2r_rxr}). For open-vocabulary object navigation on HM3D-OVON, the Diffusion Policy variant shows the strongest results with a 4.9\% absolute improvement over previous state-of-the-art, while the Action Former policy achieves a 4.3\% improvement (see Table \ref{tab:ovon}). 

\begin{figure*}[t]
    \centering
    \includegraphics[width=\textwidth]{imgs/results.pdf} 
    \caption{Qualitative results. Our model predict the next five visual frames while simultaneously outputting the corresponding 5-step waypoint sequence. The generated frames are in strong agreement with the scenes rendered from the predicted waypoints and also the predicted trajectory marked with red arrows, showing strong consistency.}
    \label{fig:wide}
\end{figure*}

\subsubsection{Ablation Study}
\begin{figure*}[t]
    \centering
    \includegraphics[width=\textwidth]{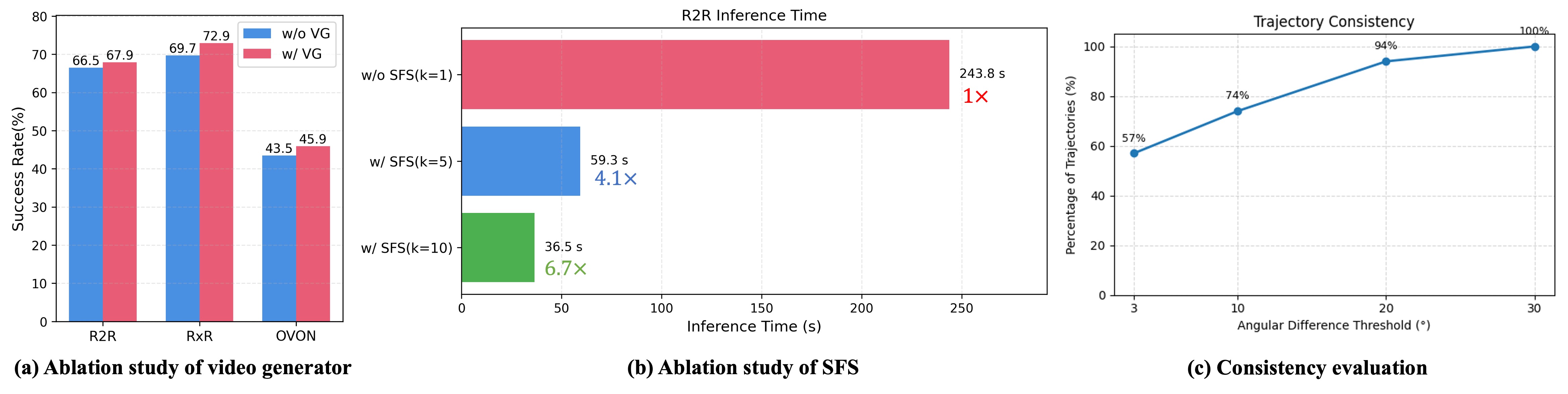} 
    \caption{\textbf{Ablation study results.} (a) \textbf{Impact of the video generator (VG):} Comparing the Success Rate (SR) with and without VG across R2R, RxR, and OVON datasets shows that predicting future observations consistently improves navigation performance. (b) \textbf{Efficiency of SFS:} Evaluation of the proposed SFS strategy on R2R with different skipping intervals $k$. The results demonstrate that increasing $k$ significantly reduces inference time (up to 6.7$\times$ speedup) while maintaining robust performance.}
    \label{fig:ab}
\end{figure*}

\noindent\textbf{Impact of the Video Generator (VG).} As shown in Fig.~\ref{fig:ab}(a), disabling the VG branch in the diffusion policy variant causes a consistent SR drop across R2R, RxR, and OVON datasets. This confirms that explicitly predicting future observations provides critical guidance for planning and improves accuracy.

\noindent\textbf{Efficiency of SFS.} Finally, we ablate the Sparse Foresight Scheduling (SFS) on 100 random R2R episodes (Fig.~\ref{fig:ab}(b)). By increasing the fixed skipping interval, inference accelerates by up to an order of magnitude while maintaining a nearly identical SR. This validates SFS as a pragmatic solution to the generation latency bottleneck.

\noindent\textbf{Scaling vs. World Modeling.} Does the performance gain stem simply from an increased parameter capacity? Table~\ref{tab:model_capacity} compares our VLA-only baseline (without VG) using 3B vs. 7B VLM backbones on R2R-CE. Scaling the VLM to 7B yields negligible improvement ($66.5\% \rightarrow 66.6\%$), hitting a performance ceiling. In contrast, our joint 3B model with the VG branch pushes the SR to 67.9\% via critical $L_{VG}$ regularization. Compared to recent state-of-the-art methods using 7B/8B LLMs, our framework maintains significant superiority, proving that the world model's bidirectional constraints yield a qualitative leap beyond pure parameter scaling.

\begin{table}[ht!]
\centering
\vspace{-5pt}
\caption{Navigation Performance (SR) vs. Parameter Scale on R2R-CE Val-Unseen. UN, SN, CN, and NA denote Uni-NaVid~\cite{zhang2024uni}, StreamVLN~\cite{wei2025streamvln}, CorrectNav~\cite{yu2025correctnav}, and NaVILA~\cite{cheng2024navila}.}
\label{tab:model_capacity}
\begin{tabular}{l ccc cc}
\toprule
\textbf{Method} & UN/SN/CN & NA & Ours-3B(w/o VG) & Ours-7B(w/o VG) & Ours-3B(w/ VG) \\ 
\midrule
LM Scale  & 7B & 8B & 3B & 7B & 3B \\
SR (\%) & 47.0$\sim$65.1 & 54.0 & 66.5 & 66.6 & \textbf{67.9} \\
\bottomrule
\end{tabular}
\vspace{-5pt}
\end{table}

\subsubsection{Consistency Analysis} 
As visually demonstrated in Figure~\ref{fig:wide}, the generated future views exhibit strong qualitative consistency with the waypoint-rendered scenes. The visual appearance and motion trends are deeply coupled with the planned trajectory during both straight and turning maneuvers, with macro-geometric cues (e.g., object contours and orientations) accurately reflecting the intended camera translation and rotation.

To quantitatively validate this alignment, we go beyond standard image fidelity metrics (i.e., PSNR and FVD) to explicitly measure spatial and geometric consistency. Specifically, we employ the open-source VGGT model to estimate the relative camera pose changes across the generated image sequence, and compare them against the ground-truth relative poses extracted from the simulator-rendered frames. We compute the angular difference $\delta_a \in [0^\circ, 360^\circ]$ between these two pose variations. As presented in Fig.~\ref{fig:ab}(c), We randomly selected 100 cases and statistically analyzed the proportion of various angular differences. The results show that our future visual predictions and actions have a very high degree of consistency.

\subsubsection{Real World Testing} 
To evaluate the real-world applicability of our approach, we conducted testing of AstraNav-World on a physical robot platform without any retraining or fine-tuning on real-world data. The testing involved several navigation tasks across real-world environment, where the robot was tasked with following natural language instructions to reach specific locations or interact with objects. Despite the significant domain gap between simulation and reality, AstraNav-World achieved good results on these real-world tasks, substantially outperforming prior methods that typically require domain adaptation (refer to the project page). 

Qualitative analysis revealed that the model's ability to generate future visual predictions enabled it to anticipate challenging transitions such as doorway passages and corner turns, significantly improving navigation robustness. This remarkable zero-shot transfer capability validates the fundamental advantage of world modeling approaches: by learning the underlying physical and navigational principles rather than simply fitting data distributions in simulation environment, our model captures essential spatial relationships and dynamics that generalize across domains. These results establish AstraNav-World as one of the first navigation models to achieve competitive real-world performance without any real-world training data, demonstrating the true potential of generative world models for embodied navigation.

\section{Conclusion}

This paper addresses core issues in embodied navigation caused by the loosely coupled "envision-then-plan" paradigm, including physical and causal inconsistency and error accumulation, by proposing a unified generative world model for navigation, AstraNav-World. It tightly binds "envision the future" and "plan the future" within a single probabilistic framework. Centered on a VLM with long-horizon understanding as the planner, the framework integrates future visual prediction with action planning through two different approaches: an Action Former policy for deterministic action generation and a diffusion policy with MMFCA for probabilistic action prediction. To handle multi-view inputs, we introduce a 3D-RoPE rearrangement method that maintains precise spatial-temporal alignment for current frames from different camera perspectives. The model is trained through a two-stage process that first develops strong individual capabilities through staged pretraining, followed by end-to-end joint optimization to enhance both executability and robustness.

Experiments show that AstraNav-World leads on the R2R-CE, RxR-CE, and HM3D-OVON navigation benchmarks with both policy variants; removing the video generation branch results in metric degradation for both implementations. Notably, our framework demonstrates exceptional generalization capabilities, with models trained exclusively in simulation achieving strong performance in real-world environments, validating the world model's ability to capture fundamental physical and navigational principles. Qualitative analysis indicates that the generated future viewpoints are highly consistent with the pose changes along the planned trajectory, markedly improving decision interpretability. To balance real-time performance, we adopt time-sliced inference so that video generation and the diffusion policy collaborate at fixed intervals, maintaining accuracy while reducing computational overhead.

AstraNav-World uses a strong VLM, physics-consistent video generation, and task-aligned policy prediction to produce coherent, executable long-horizon navigation plans. Challenges include generation latency and compute limits in complex scenes. Future work will scale to longer horizons and harder tasks, strengthen physical and causal world modeling, improve closed-loop consistency and real-time reasoning/planning, and make embodied agents more reliable and interpretable in the real world.

\clearpage
\newpage
\bibliographystyle{assets/plainnat}
\bibliography{paper}

\end{document}